\documentclass[conference]{IEEEtran}
\usepackage{cite}
\usepackage{amsmath,amssymb,amsfonts}
\usepackage{algorithmic}
\usepackage{graphicx}
\usepackage{textcomp}
\usepackage{xcolor}
\usepackage{subcaption}

\def\BibTeX{{\rm B\kern-.05em{\sc i\kern-.025em b}\kern-.08em
    T\kern-.1667em\lower.7ex\hbox{E}\kern-.125emX}}
\begin{document}

\title{Real-Time Spacecraft Pose Estimation Using Mixed-Precision Quantized Neural Network on COTS Reconfigurable MPSoC}

\author{
    \IEEEauthorblockN{
        Julien Posso\IEEEauthorrefmark{1},
        Guy Bois\IEEEauthorrefmark{1},
        Yvon Savaria\IEEEauthorrefmark{2}
    }
    \IEEEauthorblockA{
        \IEEEauthorrefmark{1}Department of Computer Engineering, \\
        \IEEEauthorrefmark{2}Department of Electrical Engineering, \\
        École Polytechnique de Montréal, Québec, Canada \\
    }
}

\maketitle
\IEEEpeerreviewmaketitle

\begin{abstract}

This article presents a pioneering approach to real-time spacecraft pose estimation, utilizing a mixed-precision quantized neural network implemented on the FPGA components of a commercially available Xilinx MPSoC, renowned for its suitability in space applications. Our co-design methodology includes a novel evaluation technique for assessing the layer-wise neural network sensitivity to quantization, facilitating an optimal balance between accuracy, latency, and FPGA resource utilization. Utilizing the FINN library, we developed a bespoke FPGA dataflow accelerator that integrates on-chip weights and activation functions to minimize latency and energy consumption. Our implementation is 7.7 times faster and 19.5 times more energy-efficient than the best-reported values in the existing spacecraft pose estimation literature. Furthermore, our contribution includes the first real-time, open-source implementation of such algorithms, marking a significant advancement in making efficient spacecraft pose estimation algorithms widely accessible. The source code is available at https://github.com/possoj/FPGA-SpacePose.

\end{abstract}

\section{Introduction}

Estimating the relative pose (position and orientation) of a known, uncooperative spacecraft from a monocular image is a crucial task in computer vision. It aims to enhance the autonomy of in-orbit spacecraft operations such as formation flying, autonomous docking, satellite maintenance, and debris removal \cite{proenca_deep_2020}. The upcoming ClearSpace-1 mission, scheduled for 2026, underscores the significance of debris removal for space sustainability and will represent the first instance of a space mission that integrates a vision-based pose estimation algorithm \cite{noauthor_clearspace_2023}.

The challenge posed by the European Space Agency (ESA) \cite{kisantal_satellite_2020} has significantly boosted interest in spacecraft pose estimation (SPE). It focuses on estimating the pose of the Tango spacecraft using the Spacecraft Pose Estimation Dataset (SPEED) \cite{sharma_pose_2019}, which has become a benchmark for training and evaluating SPE algorithms. This dataset is utilized in our study.

Two prominent techniques based on convolutional neural networks (CNNs) have emerged as highly effective in the SPE domain: keypoint detection \cite{chen_satellite_2019, kisantal_satellite_2020, black_real-time_2021, hu_wide-depth-range_2021, cassinis_leveraging_2023, legrand_end--end_2023} and a combination of soft classification with direct regression \cite{proenca_deep_2020, posso_mobile-ursonet_2022}. However, challenges persist in the area of real-time SPE inference, with only a few publications addressing this issue \cite{black_real-time_2021, cosmas_utilization_2020, wang_ca-spacenet_2022, lotti_deep_2022}.

This paper proposes a pioneering approach to real-time SPE on a commercial off-the-shelf (COTS) Xilinx MPSoC, already utilized in the space domain \cite{perez_run-time_2020}. We begin with a trained Float32 Mobile-URSONet model, featuring an embedded-friendly MobileNetV2 backbone \cite{posso_mobile-ursonet_2022}. Mixed-precision quantization is used to find an optimal balance between accuracy, latency, and FPGA resource utilization for each neural network layer. We develop a custom energy-efficient dataflow accelerator and implement it on the programmable logic available on the Xilinx MPSoC, commonly called FPGA. This co-design methodology aims to provide high-quality SPE algorithms executed on low-cost, low-power, space-ready embedded computers. The contributions of our paper include:
\begin{itemize}
    \item The first real-time inference of a popular SPE neural network on a space-ready Xilinx MPSoC. Our FPGA dataflow accelerator is 7.7 times faster and 19.5 times more energy-efficient than the best-reported values in the existing SPE literature.
    \item An in-depth study of layer-wise mixed-precision quantization for the SPE neural network. This study leads us to propose a method for selecting the bit-width of the mixed-precision neural network.
    \item The first open-source, real-time implementation of an SPE neural network \cite{posso_fpga-spacepose_2024}.
\end{itemize}

\section{Related Works}
\label{sec:related_works}

Recent advancements in spacecraft pose estimation (SPE) have been driven by keypoint detection \cite{chen_satellite_2019, kisantal_satellite_2020, hu_wide-depth-range_2021, cassinis_leveraging_2023, legrand_end--end_2023} and soft classification techniques \cite{proenca_deep_2020, posso_mobile-ursonet_2022}. Keypoint detection involves identifying and locating \textit{a priori} defined landmarks of the target within an image, and combining them with a 3D model of these landmarks to recover the pose \cite{kisantal_satellite_2020}. In contrast, soft classification assigns a probability distribution across predefined orientation classes in a discrete output space. When combined with direct regression for position estimation, this method has demonstrated the best generalization capabilities \cite{posso_mobile-ursonet_2022} and the potential to extend pose estimation to unknown targets \cite{proenca_deep_2020}. However, the neural network complexity and embeddability balance have received less attention. The works of \cite{black_real-time_2021} and \cite{posso_mobile-ursonet_2022} have significantly contributed to the embeddability of SPE algorithms by leveraging the MobileNetV2 architecture \cite{sandler_mobilenetv2_2018}. We argue that quantizing these previously reported models would further enhance their embeddability. Our paper primarily focuses on Mobile-URSONet \cite{posso_mobile-ursonet_2022}, an open-source framework, but the same methodology could be applied to the work presented in \cite{black_real-time_2021}.

Mixed-precision quantization \cite{dong_hawq_2019, wang_haq_2019, uhlich_mixed_2020, askarihemmat_barvinn_2023, tang_mixed-precision_2023, rakka_mixed-precision_2022} has emerged as a superior method for enhancing neural networks' latency and energy efficiency beyond uniform Int8 quantization, primarily applied to benchmark datasets like CIFAR, ImageNet, and VOC. Uniquely, \cite{javed_module-wise_2023} introduced a module-wise quantization tailored for SPE but did not achieve real-time inference on embedded computers.

Previous research on real-time embedded SPE has made significant strides, as seen in works by \cite{black_real-time_2021}, \cite{lotti_deep_2022}, \cite{cosmas_utilization_2020}, and \cite{wang_ca-spacenet_2022}, employing various hardware and quantization strategies. However, these studies often lack detailed methodology, particularly regarding quantization's effects on accuracy and efficiency, and frequently omit source code, hindering reproducibility. The absence of exhaustive performance metrics such as throughput and power consumption also restricts the understanding of their practical efficiency and applicability. Our research addresses these gaps by prioritizing methodological transparency, reproducibility, and comprehensive performance evaluation.

\section{Co-Design Methodology}

\subsection{Methodology Overview}

\begin{figure}[ht]
    \includegraphics[scale=0.11]{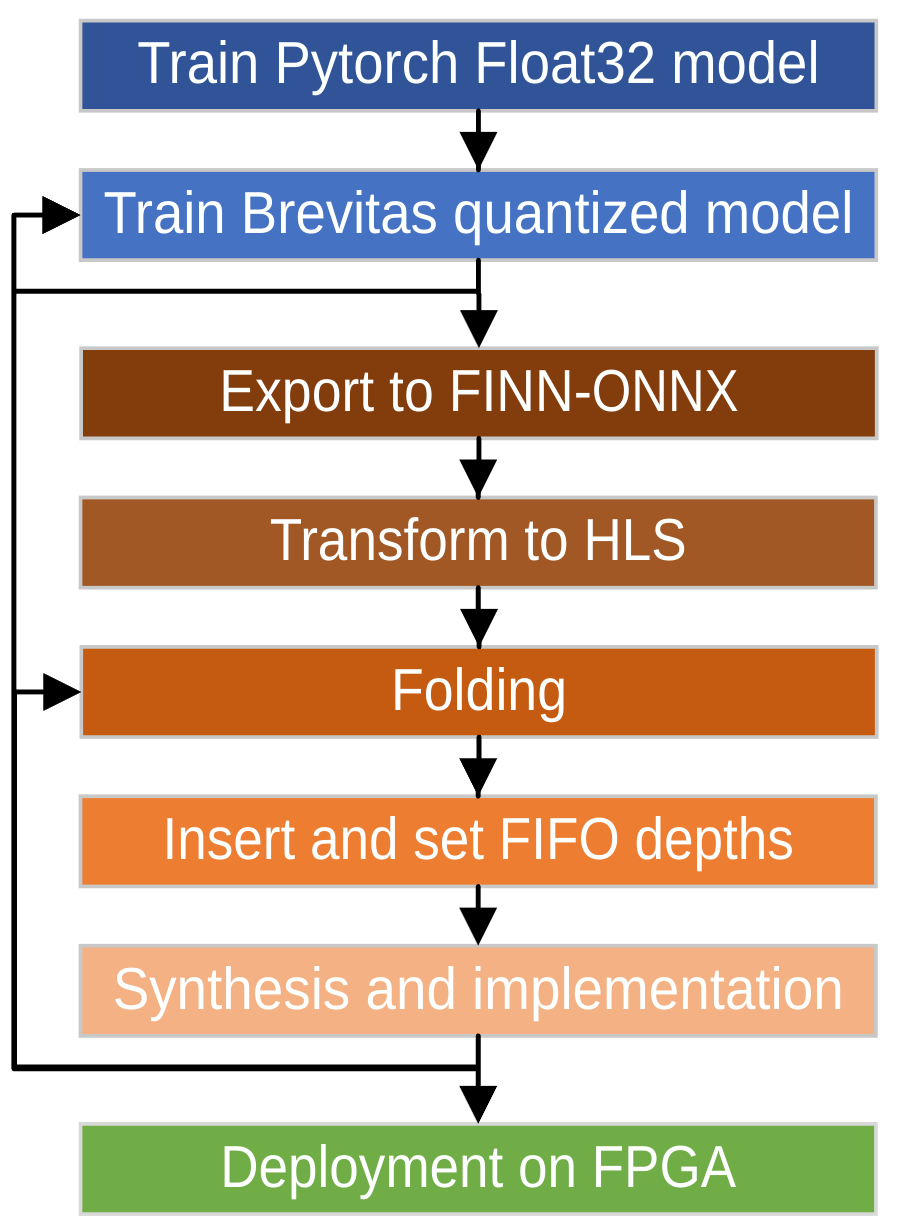}
    \centering
    \caption{Methodology that targets FPGA inference}
    \label{fig:fpga_flow}
\end{figure}

Figure \ref{fig:fpga_flow} outlines our proposed co-design methodology, which achieves an optimal balance between accuracy, latency, and FPGA resource utilization. The process begins with the Float32 weights of Mobile-URSONet \cite{posso_mobile-ursonet_2022}. Using Brevitas \cite{pappalardo_xilinxbrevitas_2021}, we quantize the network for Quantization-Aware Training (QAT) employing per-channel symmetric uniform quantization with layer-wise arbitrary precision. Over 20 epochs, we trained 146 model variants to optimize bit-widths across layers, a process detailed in Section \ref{sec:mixed-precision}. Although such an exhaustive approach would be impractical for large datasets like ImageNet \cite{dong_hawq_2019, wang_haq_2019}, it was feasible within a week for a smaller dataset using an Nvidia P100 GPU. This efficiency was partly due to acceleration from Float32 pre-training, which reduced QAT time by a factor of three. Subsequently, we converted the neural network into a FINN-ONNX graph, which was then transformed into FINN C++ High-Level Synthesis (HLS)-compatible nodes \cite{noauthor_finn-hlslib_2023}, ensuring a one-to-one correspondence between the nodes and the neural network layers, including im2col and matrix-vector multiplication nodes for convolutions. Our custom folding algorithm was designed to optimize node parallelism, minimizing FPGA resource usage while meeting our target latency constraints, in line with FINN’s inter-layer constraints. Through Python simulations, we fine-tuned FIFO depths to ensure optimal data flow, a process elaborated in Section \ref{sec:finn_exp-results}. Post-synthesis, meeting FPGA resource constraints may require design optimizations, such as reducing parallelism or adjusting the QAT bit-widths. The final outcome is an accelerator pipeline, with one accelerator per CNN layer, featuring on-chip weights and activations to minimize both energy use \cite{horowitz_11_2014} and latency, compared to conventional accelerators \cite{blott_finn-_2018}.

\subsection{Neural Network Architecture}

\begin{figure}[ht]
    \includegraphics[scale=0.09]{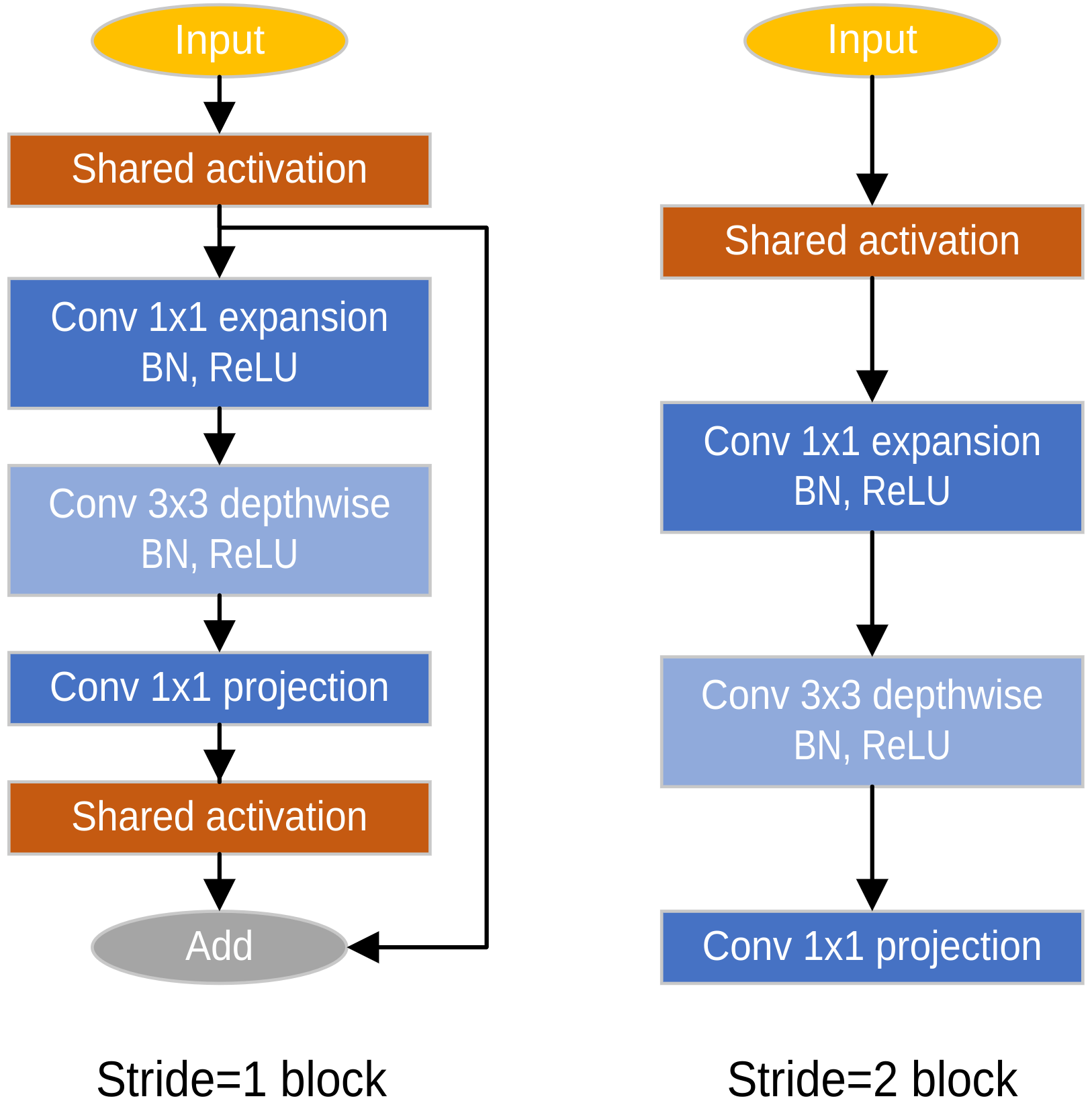}
    \centering
    \caption{Modified inverted residual block}
    \label{fig:inverted_residual_archi}
\end{figure}

Mobile-URSONet leverages a MobileNetV2 backbone \cite{posso_mobile-ursonet_2022}, beginning with a 3x3 convolutional layer, followed by 17 inverted residual blocks, and ending with a 1x1 convolutional layer. We have tailored the architecture of the inverted residual blocks, as illustrated in Figure \ref{fig:inverted_residual_archi}, by integrating linear activations with shared scaling factors. This adaptation enables quantized tensor summation and allows for entirely integer-based computations throughout the network after applying FINN transformations. Moreover, we maintain true residual connections in the neural network, contrary to the FINN authors who insert activation functions or convolutions in every shortcut to facilitate the transformation to HLS-compatible nodes at the cost of the network's simplicity \cite{noauthor_finn_nodate}. While the parameter count of MobileNetV2 increases across its layers, the number of Multiply-Accumulate (MAC) operations remains comparatively consistent. This stability is attributed to the feature maps' diminishing dimensions (width and height) as the network deepens, counterbalancing the increase in parameter count.

\subsection{Mixed-Precision Quantization}
\label{sec:mixed-precision}

\begin{figure*}[ht]
     \centering
     \begin{subfigure}[b]{0.3\textwidth}
         \centering
         \includegraphics[width=\textwidth]{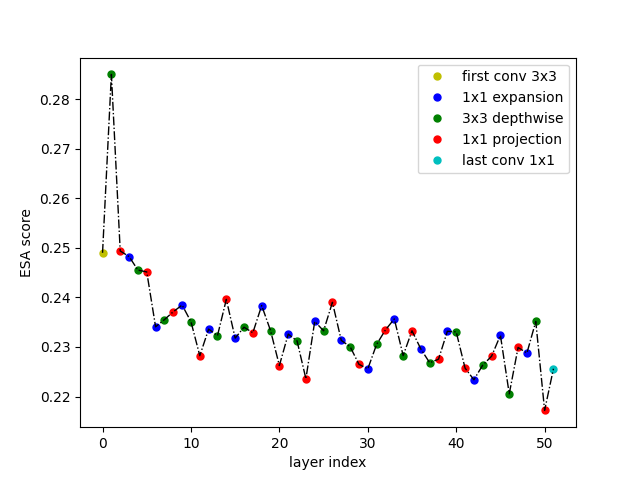}
         \caption{ESA score vs. weight layer index binarization}
         \label{fig:esa_score_vs_layer_index}
     \end{subfigure}
     \hfill
     \begin{subfigure}[b]{0.3\textwidth}
         \centering
         \includegraphics[width=\textwidth]{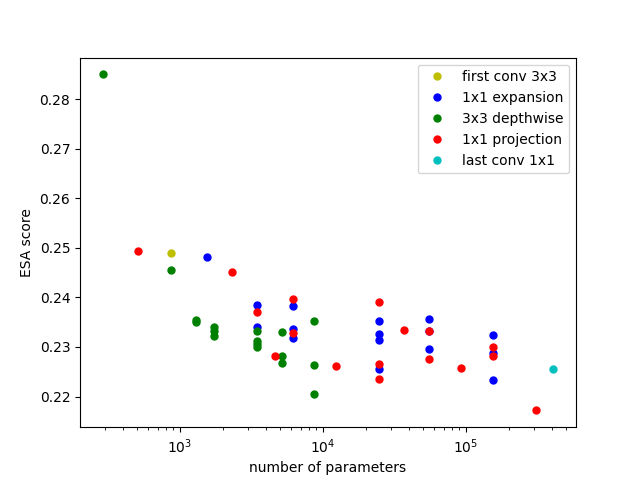}
         \caption{ESA score vs. the number of parameters in the binarized layer}
         \label{fig:esa_score_vs_params}
     \end{subfigure}
     \hfill
     \begin{subfigure}[b]{0.3\textwidth}
         \centering
         \includegraphics[width=\textwidth]{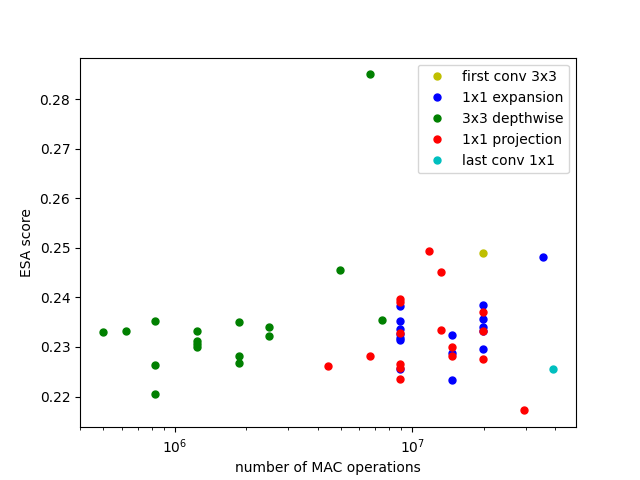}
         \caption{ESA score vs. the number of MAC operations in the binarized layer}
         \label{fig:esa_score_vs_mac_ops}
     \end{subfigure}
        \caption{ESA score when only one convolutional layer's weights are binarized, with all other layers and activations at 8 bits}
        \label{fig:backbone_weights}
\end{figure*}

This section proposes a method to assess each layer's sensitivity to low-bit quantization by setting all layers to 8-bit weights and activations, except for one layer's binarized weights. In Figure \ref{fig:esa_score_vs_layer_index}, we observe that layers closer to the input of the neural network generally exhibit greater sensitivity to quantization. This trend is consistent with Figure \ref{fig:esa_score_vs_params}, which shows that sensitivity to low-bit quantization decreases with the parameter count of a layer, and the network's parameter count grows across its layers. The considerable volume of MAC operations in the early layers, driven by high-dimensional feature maps, does not reduce the sensitivity to quantization, as shown in Figure \ref{fig:esa_score_vs_mac_ops}. These findings challenge the prevailing belief that quantization impacts the first and last layers most significantly \cite{liu_layer_2021, wang_haq_2019}. Notably, the second convolutional layer diverges from these expectations \cite{wang_haq_2019}, a variance linked to a characteristic of MobileNetV2. Specifically, the architecture's first inverted residual block omits the expansion layer to keep a reasonable MAC operations count in the subsequent layer. This results in the initial depthwise convolution having merely 288 parameters, as highlighted in Figure \ref{fig:backbone_weights}. This finding suggests that architectural modifications in recent designs \cite{woo_convnext_2023}, influenced by vision transformers that position depthwise convolutions before expansion layers, might be sensitive to low-bit quantization.
Further experiments have revealed that activations are notably more sensitive to low-bit quantization, with quantization errors below 4 bits frequently causing orientation decoding errors and disrupting the neural network's training process. This significant constraint influences the layer-wise bit-width activation choices in the next paragraph. Our findings also indicate that the binarization of early layers diminishes the generalization capabilities of the neural network. However, it improves in later layers, suggesting that low-bit quantization compromises information in the initial layers while acting as regularization in the last layers. Additionally, the sensitivity of early layers to quantization predominantly affects orientation estimation, as opposed to position estimation, which is comparatively simpler \cite{posso_mobile-ursonet_2022}. Therefore, quantization below 8 bits has a more pronounced impact on orientation accuracy than on position accuracy.

This analysis leads us to identify an optimal balance among layer-wise bit-width, FPGA resource utilization, and throughput. Given the limitations of FINN, we quantize only the neural network's backbone for FPGA deployment, leaving the heads in Float32. We observed that activations are highly sensitive to low-bit quantization, yet the hardware resources required increase exponentially with the activation bit-width due to the current FINN C++ HLS backend \cite{noauthor_finn-hlslib_2023}. Consequently, we quantize all activations to four bits to conserve FPGA resources, as higher bit-widths would disproportionately consume resources \cite{blott_finn-_2018}. Weight bit-width selection varies by layer index due to differing quantization sensitivities; the weights of the first layer are quantized at four bits, reflecting their higher sensitivity. The most sensitive layer, the first depthwise convolution, uses six-bit weights. Subsequently, the first projection convolution weights are set to four bits, and the weights of all remaining 49 convolutional layers are set to three bits. This configuration will be used in the next section to evaluate and implement the neural network on the FPGA.

\subsection{Experimental Results}
\label{sec:finn_exp-results}

Table \ref{tab:fpga_score} presents a comparison of pose estimation metrics across the FPGA workflow, including ESA score (where lower is better), position error (\(e_t\)), and orientation error (\(e_q\)) \cite{kisantal_satellite_2020}, from initial Float32 inferences on a computer to FPGA deployment via FINN, with images sized at 240x240 to meet FINN's requirements. The shift to Int8 quantization leads to a slight increase in the ESA score, primarily attributed to a rise in position error. This is likely due to the position estimation's reliance on direct regression, which is more susceptible to optimization effects, in contrast to orientation estimation, which employs a probabilistic method through soft classification and thus exhibits greater robustness. Additionally, the quantization process introduces a regularization effect that benefits orientation estimation, given the large number of parameters in its neural network branch. Mixed-precision quantization affects both position and orientation errors despite using larger bit-widths for the most sensitive CNN layers, as explained in Section \ref{sec:mixed-precision}. Nevertheless, the resulting degradation is kept to a manageable level. Our CNN FPGA accelerator generates outputs that exactly match the tensors computed with Brevitas on the computer. The mean square error analysis reveals no differences, ensuring identical Pose metrics on both the FPGA and computer, underscoring our FPGA accelerator's meticulous design and validation.

\begin{table}[ht]
    \centering
    \caption{Pose metrics from the Float32 on computer inference to the onboard implementation on the SPEED dataset}
    \begin{tabular}{|l|c|c|c|}
        \hline
        Model     & ESA score $\downarrow$ & $e_q$ (°) & $e_t$ (m) \\
        \hline
        Float32         & 0.296      & 13.0      & 0.71     \\
        Int8            & 0.307      & 13.2      & 0.80     \\
        mixed-precision & 0.411      & 18.7      & 0.91     \\
        FPGA on-board   & 0.411      & 18.7      & 0.91     \\ 
        \hline
    \end{tabular}
    \label{tab:fpga_score}
\end{table}

Our FPGA accelerator operates at 187.5 MHz and utilizes 91\% of the Lookup Tables (LUTs) and 90\% of the Block RAMs (BRAMs), but only 32\% of the Digital Signal Processors and 26\% of the flip-flops on the FPGA. Contrary to the expectations of the FINN library authors \cite{blott_finn-_2018}, computing activation functions significantly impact FPGA resources, as it requires as many LUTs as computing convolutions. This discrepancy could stem from using the MobileNetV2 architecture tailored for mobile inference, unlike the less efficient ResNet architecture employed by FINN's authors. The multi-threshold units and convolution computations primarily consume flip-flops and LUTs. At the same time, nearly all BRAMs are dedicated to FIFOs between units, indicating potential for optimization in FINN's resource allocation strategies.

Table \ref{tab:board_finn} presents FPGA implementation results compared to estimates, covering power consumption (static and dynamic), frames per second (FPS), and energy efficiency. Our accelerator is structured as a pipeline of units connected by FIFOs, with its throughput limited by the slowest unit, which operates at 250 FPS. Combined with the power consumption derived from the Xilinx Vivado post-implementation report, it indicates high energy efficiency, as detailed in Section \ref{sec:summary}. On-board testing reveals a significant throughput discrepancy, more than four times lower than estimated, attributed to insufficient BRAMs, which lead to undersized FIFOs. This constraint slows down the pipeline and reduces power consumption by a factor of 4.4. Despite this limitation, the results underscore the real-time energy-efficient capabilities of our FPGA-accelerated SPE algorithm, with further comparisons provided in Section \ref{sec:summary}.


\begin{table}[ht]
    \centering
    \caption{FPGA implementation metrics}
    \begin{tabular}{|l|c|c|c|}
        \hline
        Board & Power (W) & FPS & FPS per Watt \\
        \hline
        Estimation & 3.83 & 250 & 65.3 \\
        ZCU104 & 0.865  & 58.7 & 67.9 \\
        \hline
    \end{tabular}
    \label{tab:board_finn}
\end{table}

\section{Comparison}

\subsection{Results Summary}
\label{sec:summary}

Table \ref{tab:summary_results} summarizes the performance of our implementations compared to existing works. The implementation by \cite{cosmas_utilization_2020} does not provide throughput or latency data, which prevents the computation of the energy efficiency of their implementation. Given that they use a Xilinx Ultra96 board, which contains a single-core Xilinx DPU on a small FPGA, achieving real-time inference may not be possible. The other FPGA Ultra96 implementation by \cite{wang_ca-spacenet_2022} implemented only a single layer of their neural network on the FPGA, also hindering the calculation of energy efficiency. Our implementation is almost 9 times faster and over 38 times more energy efficient than the Intel Atom-based implementation by \cite{black_real-time_2021}. Additionally, our FPGA solution is 7.7 times faster and 19.5 times more energy efficient than the Google Edge TPU-based ASIC by \cite{lotti_deep_2022}, highlighting the efficiency of our mixed-precision dataflow FPGA accelerator.

\begin{table}[ht] 
    \centering
    \caption{Comparison with the existing literature}
    \begin{tabular}{|l|c|c|c|}
        \hline
        Implementation & Power (W) & FPS & FPS per Watt \\
        \hline
        FPGA ZCU104 (ours)  & 0.865 & 58.7 & 67.9 \\
        CPU Intel Atom \cite{black_real-time_2021} & 3.7 & 6.58 & 1.78 \\
        FPGA Ultra96 \cite{cosmas_utilization_2020} & 1.32 & x & x \\
        FPGA Ultra96 \cite{wang_ca-spacenet_2022} & x & 167 & x \\
        Google edge TPU \cite{lotti_deep_2022} & 2.2 & 7.66 & 3.48 \\
        \hline
    \end{tabular}
    \label{tab:summary_results}
\end{table}

\subsection{Limitations and Future Works}

This work advances embedded in-orbit SPE, highlighting achievements and suggesting future directions. Integrating the neural network's backbone into the FPGA marks a significant step forward, with potential enhancements including transitioning the network's head from the CPU to the FPGA for full capability utilization. Future research could explore additional SPE neural networks, such as \cite{black_real-time_2021}'s keypoint-based model, despite challenges in experiment reproducibility due to unavailable code and data. Investigating newer convolutional architectures such as ConvNext V2 \cite{woo_convnext_2023} could offer accuracy improvements and insights into the impact of low-bit quantization on advanced model structures, particularly regarding quantization effects on modified inverted residual blocks. Further experiments are essential to validate these preliminary findings.

\section{Conclusion}

This paper aims to provide efficient spacecraft pose estimation (SPE) algorithms that can be effectively implemented on low-cost, low-power, space-ready embedded computers to achieve autonomous space navigation. We propose a co-design methodology to achieve real-time and energy-efficient inference of a popular SPE neural network on the FPGA of a commercial, space-proven Xilinx MPSoC. Thanks to the use of mixed-precision quantization and custom dataflow FPGA acceleration, our implementation is 7.7 times faster and 19.5 times more energy-efficient than the best previously reported SPE algorithms. Furthermore, we present the first real-time, open-source implementation, marking a significant step toward democratizing efficient spacecraft pose estimation algorithms.

\section{Acknowledgments}

The authors thank the Canadian Space Agency, MITACS, and Space Codesign Systems for their financial support.

\bibliography{references}
\bibliographystyle{IEEEtran}

\end{document}